\documentclass[11pt]{article}
\pdfoutput=1 
\usepackage[final]{acl}

\usepackage{times}
\usepackage{latexsym}

\usepackage[T1]{fontenc}

\usepackage[utf8]{inputenc}
\usepackage{amsthm}

\usepackage{microtype}

\usepackage{inconsolata}

\usepackage{graphicx}

\usepackage{multirow}
\usepackage{booktabs}
\usepackage{arydshln}  
\usepackage{amsmath}  %
\usepackage{amssymb} %
\usepackage{graphicx} 
\usepackage[table]{xcolor} 
\usepackage{graphicx}
\usepackage{xcolor}      
\usepackage[most]{tcolorbox}  
\usepackage[normalem]{ulem}
\usepackage{float} 
\usepackage{algorithm}
\usepackage{algorithmic}

\usepackage{xcolor,pifont}
\newcommand{\cmark}{\textcolor{green}{\ding{51}}}  
\newcommand{\xmark}{\textcolor{red}{\ding{55}}}   

\definecolor{blue1}{rgb}{0.118, 0.251, 0.486}
\definecolor{blue2}{rgb}{0.918, 0.949, 0.996}
\definecolor{lightpurple}{rgb}{0.941, 0.941, 1.000} 
\definecolor{darkpurple3}{rgb}{0.35, 0.35, 0.80}
\definecolor{darkpurple}{rgb}{0.25, 0.25, 0.60}   
\definecolor{normalpurple}{rgb}{0.62, 0.436, 0.706}   

\definecolor{backcolour}{rgb}{0.95,0.95,0.92}
\newtcolorbox[auto counter]{promptbox}[2][]{
  enhanced,                 
  colback=lightpurple,
  colframe=black,
  fontupper=\normalsize,         
  fonttitle=\bfseries\normalsize,
  title=Prompt~\thetcbcounter: #2,
  #1
}

\definecolor{darkblue}{rgb}{0, 0, 0.5}
\hypersetup{colorlinks=true, citecolor=darkblue, linkcolor=darkblue, urlcolor=darkblue}

\newcommand{\think}[1]{\textcolor{blue}{\texttt{<think>}} #1 \textcolor{blue}{\texttt{</think>}}}
\newcommand{\search}[1]{\textcolor{cyan}{\texttt{<search>}} #1 \textcolor{cyan}{\texttt{</search>}}}
\newcommand{\info}[1]{\textcolor{brown}{\texttt{<information>}} #1 \textcolor{brown}{\texttt{</information>}}}
\newcommand{\answer}[1]{\textcolor{purple}{\texttt{<answer>}} #1 \textcolor{purple}{\texttt{</answer>}}}

\title{CriticSearch: Fine-Grained Credit Assignment for Search Agents via a Retrospective Critic}

\author{
  Yaocheng Zhang$^{1,2}$\thanks{Equal contribution.} \quad
  Haohuan Huang$^{1,3}$\footnotemark[1] \quad
  Zijun Song$^{1,2}$ \quad
  Yuanheng Zhu$^{1,3}$\thanks{Corresponding author.} \\
  \textbf{Qichao Zhang}$^{1,3}$ \quad
  \textbf{Zijie Zhao}$^{1,3}$ \quad
  \textbf{Dongbin Zhao}$^{1,3}$\\
 $^{1}$ Institute of Automation, CAS, Beijing, CHINA \\
  $^{2}$School of Advanced Interdisciplinary Sciences, UCAS, Beijing, CHINA\\
 $^{3}$School of Artificial Intelligence, UCAS, Beijing, CHINA \\
  \texttt{zhangyaocheng2023@ia.ac.cn}
}

\begin{document}
\maketitle
\begin{abstract}
Tool-Integrated Reasoning (TIR) with search engines enables large language models to iteratively retrieve up-to-date external knowledge, enhancing adaptability and generalization in complex question-answering tasks.   However, existing search agent pipelines typically depend on reinforcement learning based optimization, which often suffers from sparse outcome rewards, leading to inefficient exploration and unstable training.     We introduce CriticSearch, a fine-grained credit-assignment framework that supplies dense, turn-level feedback via a retrospective critic mechanism.   During training, a frozen, asymmetric critique LLM retrospectively evaluates each turn using privileged information from the full trajectory and gold answers, converting these assessments into stable, dense rewards that guide policy improvement.     Experimental results across diverse multi-hop reasoning benchmarks demonstrate that CriticSearch consistently outperforms existing baselines, achieving faster convergence, improved training stability, and higher performance.

\end{abstract}

\section{Introduction}

\begin{table*}[t]
\small
\centering
\begin{tabular}{cccc}
\toprule
\textbf{Method} &\textbf{\shortstack{Dense  Reward}}& \textbf{\shortstack{No additional Rollout}} & \textbf{\shortstack{Generality}} \\
\midrule
Search-R1 \citep{jin2025searchr1trainingllmsreason} & \xmark& \cmark  & \cmark \\
StepSearch \citep{wang2025stepsearchignitingllmssearch}& \cmark & \cmark &\xmark \\
ARPO \citep{dong2025agenticreinforcedpolicyoptimization} &\cmark& \xmark & \cmark \\
ReasonRAG \citep{zhang2025processvsoutcomereward}&\cmark& \xmark & \cmark \\
CriticSearch (Ours) & \cmark & \cmark &\cmark \\
\bottomrule
\end{tabular}
\caption{\textbf{Comparison of agentic RL with search engines.} CriticSearch satisfies three criteria:
1) dense reward, 2) no need for additional rollout during training, and 3) generality to all training data.}
\label{tab:compare_method}
\end{table*}

Recently, multi-turn Tool-Integrated Reasoning (TIR) has achieved great potential across a wide range of downstream tasks, including code generation, mathematical reasoning, and question-answering (Q\&A) \citep{shao2024deepseekmathpushinglimitsmathematical,jin2025searchr1trainingllmsreason,feng2025retoolreinforcementlearningstrategic}.     This paradigm draws inspiration from the interaction and iterative mechanisms in reinforcement learning (RL) \citep{schulman2017proximalpolicyoptimizationalgorithms,shao2024deepseekmathpushinglimitsmathematical}, enabling models to invoke external tools (e.g., search engines, code interpreters), access up-to-date information, and progressively refine their reasoning process through feedback-driven iterations \citep{jin2025searchr1trainingllmsreason,feng2025retoolreinforcementlearningstrategic}.    Compared with traditional single-turn reasoning, TIR effectively mitigates inherent limitations of large language models (LLMs), such as outdated knowledge and insufficient contextual information, thereby achieving stronger performance and better generalization in complex Q\&A and problem-solving scenarios.

\begin{figure}[!t]
    \centering
    \includegraphics[width=0.8\linewidth]{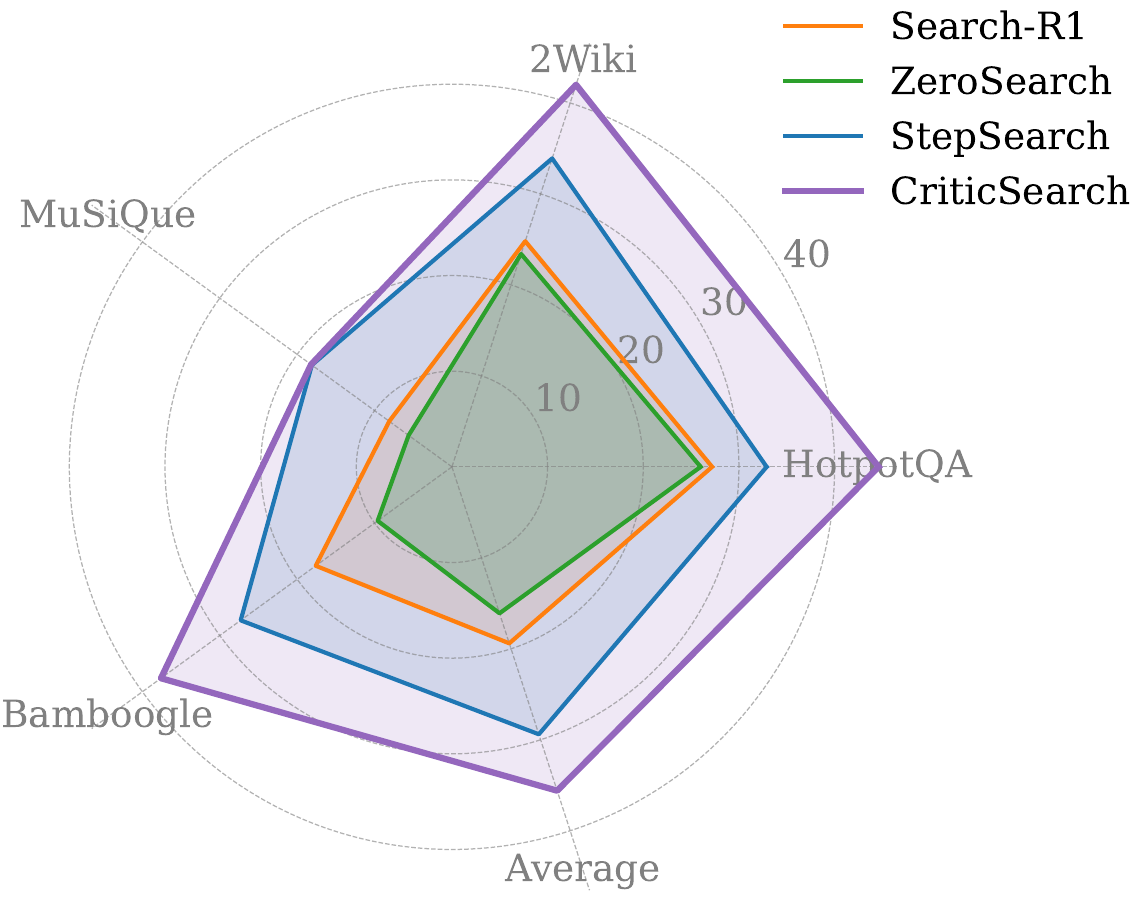}
    \caption{CriticSearch achieves leading performance on most of the datasets compared with RL methods.}
    \label{fig:ladar_chart}
\end{figure}

However, the effectiveness of TIR hinges on RL-based optimization.      Existing algorithms typically rely on sparse outcome rewards, where all actions within a trajectory share the same reward regardless of quality \citep{jin2025searchr1trainingllmsreason,feng2025retoolreinforcementlearningstrategic,chen2025researchlearningreasonsearch,zheng2025deepresearcherscalingdeepresearch}.      Such coarse feedback prevents the model from distinguishing between effective and ineffective tool-calling, leading to inefficient exploration and slow policy improvement \citep{wang2025stepsearchignitingllmssearch,dong2025agenticreinforcedpolicyoptimization}.      Moreover, the reward sparsity induces high variance in returns, making training unstable \citep{plappert2018multigoalreinforcementlearningchallenging,zyc2025aamas}.      These challenges become particularly severe in deep-search (search agent) or frequent tool-call settings, where a large number of externally generated tokens exacerbate instability, often resulting in reward collapse and gradient explosion \citep{xue2025simpletirendtoendreinforcementlearning}.

A natural solution to these challenges is to provide fine-grained, turn-level (action-level) feedback rather than relying solely on sparse global rewards \citep{feng2025groupingrouppolicyoptimizationllm,dong2025agenticreinforcedpolicyoptimization, wang2025stepsearchignitingllmssearch}.
However, existing search agent paradigms struggle to perform effective credit assignment under sparse rewards (Tab.~\ref{tab:compare_method}). \citet{feng2025groupingrouppolicyoptimizationllm,dong2025agenticreinforcedpolicyoptimization} use Monte Carlo to approximate fine-grained rewards, which often suffer from high variance and require large amounts of rollout for stable estimation. 
Some studies have attempted to use ground-truth references for supervision, but such methods depend heavily on annotated data and lack generality \citep{wang2025stepsearchignitingllmssearch}.

To overcome these limitations, we propose \textit{CriticSearch}—a fine-grained credit assignment method that leverages LLM to assist policy training. 
The key idea of CriticSearch is to introduce a \emph{retrospective critic} mechanism: after the model produces a complete reasoning trajectory and obtains the gold answer, a critique model revisits each intermediate action to assess its contribution toward the final outcome and generate fine-grained turn-level rewards. Compared with forward-looking decision-making, this retrospective assessment leverages privileged information that is asymmetrically available during training (e.g., the gold answer and future interaction turns). Such hindsight access makes it easier and more reliable to generate turn-level rewards (Section~\ref{sec:accuracy}) \citep{zhou2025sweetrltrainingmultiturnllm,lee2020learning,lowe2017multi}.
By integrating both turn-level and global rewards, CriticSearch assigns lower rewards to redundant or low-quality actions while amplifying the learning signals of key actions. 
With the aid of stable and denser feedback, CriticSearch effectively enhances training stability and learning efficiency.

Our main contributions are as follows:
\begin{itemize}
    \item We propose \textit{retrospective critic}, a novel mechanism for Search Agent that uses a pre-trained LLM, without fine-tuning, to generate dense reward, providing more effective feedback.

    \item We provide valuable experimental analyses that reveal key phenomena arising from the introduction of dense rewards, including accelerated convergence and mitigation of training instability.
    \item CriticSearch surpasses dense-reward baselines by 16.7\% and 6.7\% on 3B and 7B models, respectively, across Q\&A benchmarks.

\end{itemize}

\begin{promptbox}[label=box: training prompt]{Training Prompt}
Answer the given question. \
You must conduct reasoning inside \think{and} first every time you get new information. \
After reasoning, if you find you lack some knowledge, you can call a search engine by \search{query}, and it will return the top searched results between \info{and}. \
You can search as many times as you want. \
If you find no further external knowledge needed, you can directly provide the answer inside \answer{and} without detailed illustrations. For example, \answer{Beijing}. Question:
\end{promptbox}

\begin{figure*}
    \centering
    \includegraphics[width=1\linewidth]{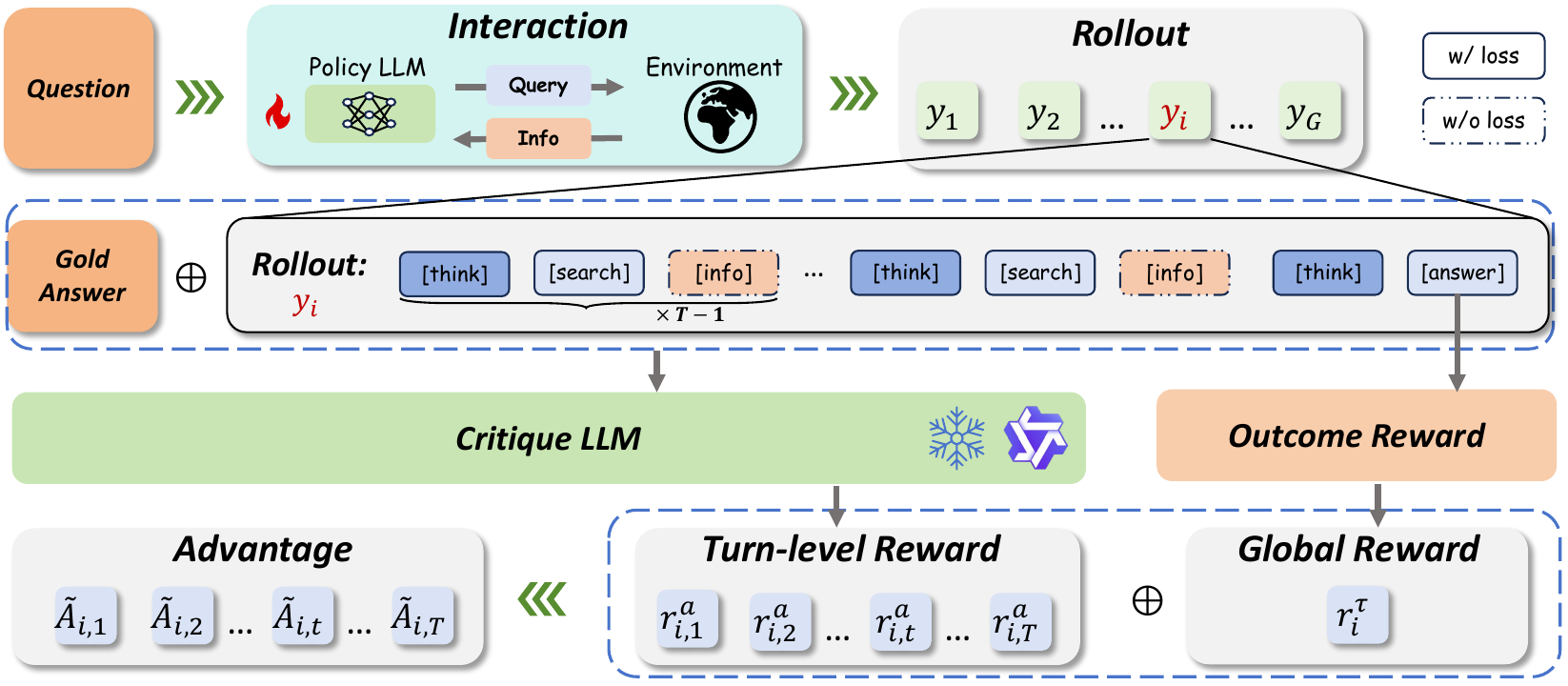}
    \caption{\textbf{Overview of CriticSearch}. The policy LLM interacts with external tools during multi-turn reasoning to generate a rollout trajectory. A frozen LLM with privileged information retrospectively evaluates each action, producing dense rewards that complement the sparse outcome reward. The resulting hybrid advantage signal provides fine-grained feedback, effectively mitigating reward sparsity in agentic RL.} 
    \label{fig:framework}
\end{figure*}

\section{Preliminaries}
\subsection{Agentic RL with Search Engines}
In a deep search task, the policy model is designed to interact with the search engine, producing a multi-turn reasoning trajectory $y = \{s_0, s_1, \ldots, s_T\}$ to answer the question $q$. Here, $T$ represents the total number of interaction turns. 
This process can be modeled as a partially observable Markov decision process (POMDP). 
As illustrated in Prompt \ref{box: training prompt}, each interaction turn $s_t = (a_t, c_t)$ consists of two stages: a \textit{decision stage} $a_t$ and a \textit{environment feedback stage} $c_t$ \citep{jin2025searchr1trainingllmsreason}. During the decision stage, the policy model performs deep reasoning and then generates an action token sequence $a_t$, which can be either a search action or an answer action. When additional information is required to answer the question, the model outputs a search action and places the corresponding search query within the \texttt{<search>...</search>} tag. 
Once the model determines that sufficient evidence has been collected, it outputs an answer action and places the final answer token $o$ within the \texttt{<answer>...</answer>} tag. In the feedback stage, the external search engine returns information based on the action generated by the policy model. 
If the model generates a search action, the search engine returns the retrieved results enclosed within the \texttt{<information>...</information>} tags, which serve as the feedback tokens $c_t$. 
When the model outputs an answer action, the search engine returns None, and the interaction process terminates.

\subsection{Group Relative Policy Optimization}
The core idea of Group Relative Policy Optimization
(GRPO) \citep{shao2024deepseekmathpushinglimitsmathematical} is to estimate the baseline through a relative reward within a group of rollouts. This approach avoids the need for a value function by computing the relative advantage of each sample within a group. Given a question $q$ and $G$ responses $\{y_i\}_{i=1}^{G}$ sampled from the old policy $\pi_{\theta_{\text{old}}}$, the GRPO objective is:

{\small
\begin{equation}
\begin{aligned}
\mathcal{J}_{\text{GRPO}}(\theta) = \mathbb{E}_{q,\{y_i\} \sim \pi_{\text{old}}} \Bigg[ 
\frac{1}{G} \sum_{i=1}^G\sum_{t=1}^{|y_i|} \mathcal{L}_{i,t} 
- \beta \, \mathbb{D}_{\text{KL}}(\pi_\theta \| \pi_{\text{ref}}) 
\Bigg]
\end{aligned}
\label{Eq:optim-obj}
\end{equation}}
where the group-wise clipped loss is defined as:
\begin{equation}
\small
\mathcal{L}_{i,t} = \min \left( w_{i,t} A_{i,t},\ \operatorname{clip}(w_{i,t},\ 1 - \epsilon,\ 1 + \epsilon) A_{i,t} \right)
\end{equation}
and the importance weight $w_{i,t}$ and normalized advantage $A_{i,t}$ are given by:
\begin{align}
w_i &= \frac{\pi_\theta(y_{i,t} \mid q, y_{i,<t})}{\pi_{\theta_{\text{old}}}(y_{i,t} \mid q, y_{i,<t})},  &A_{i,t}= \frac{r_i - \operatorname{mean}(r)}{\operatorname{std}(r)}
\end{align}
where $r_i$ is the reward assigned to response $y_i$, and $\beta$ and $\epsilon$ are tunable hyperparameters. During the training, we focus on learning the ability of deep reasoning and search. Gradients are propagated only through tokens generated by the policy model (i.e., reasoning, search query, and final answer), while the information returned by the search engine is masked out to prevent gradient propagation.

\section{Methods}
The design of reward signals is critical for improving the learning efficiency and overall performance of agentic RL \citep{wang2025stepsearchignitingllmssearch,dong2025agenticreinforcedpolicyoptimization,plappert2018multigoalreinforcementlearningchallenging,zyc2025aamas}.
In our framework, we combine global and turn-level rewards for advantage estimation in GRPO.
The global reward, determined by the correctness of the final answer, measures the overall quality of the entire reasoning trajectory, while the turn-level rewards, assigned after each search action, assess the local quality of individual reasoning steps and search queries.
The following section provides a detailed description of both reward schemes.

\begin{figure}[!t]
    \centering
    \includegraphics[width=\linewidth]{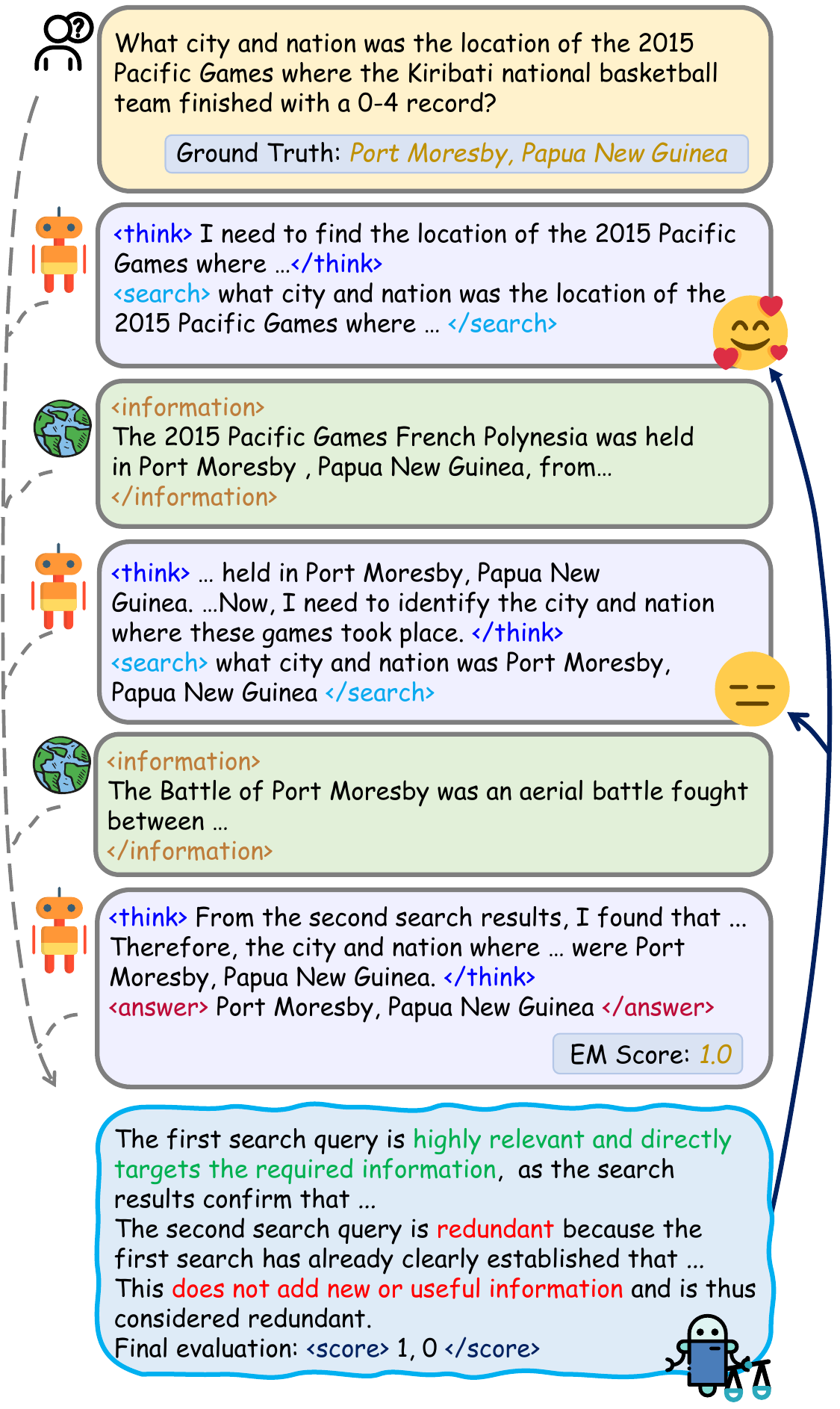}
    \caption{An example from CriticSearch illustrating the evaluation process of the Critique model. In this trajectory, the final answer is correct but involves redundant search actions. A frozen critique LLM, after deliberation, provides accurate binary rewards for each turn.
\label{fig:reard_case}
}
    \label{fig:intro_pic}
\end{figure}
 
\subsection{Global Reward Signal}
Following prior work \cite{jin2025searchr1trainingllmsreason}, in addition to employing the rule-based outcome reward function, we incorporate a format reward to construct the final reward function $r_{\phi}(q, y)$.
\begin{gather}
\small
\scalebox{1}{
$r_{\phi} = 
\begin{cases}
    1              & \text{if } o_{\text{pred}} = o_{\text{gold}} \land f_{\text{format}}(y), \\
    1 - \lambda_f  & \text{if } o_{\text{pred}} = o_{\text{gold}} \land \lnot f_{\text{format}}(y) , \\
    \lambda_f      & \text{if } o_{\text{pred}} \neq o_{\text{gold}} \land\textit{} f_{\text{format}}(y), \\
    0              & \text{if } o_{\text{pred}} \neq o_{\text{gold}} \land \lnot f_{\text{format}}(y), \\
\end{cases}$}
\end{gather}
where $f_\text{format}$ verifies whether the response $y$ follows the correct reasoning format and properly uses the special tokens (e.g., <search>...</search>).

For each question $x$, a set of responses $\{y_i\}_{i=1}^{G}$ is sampled from the policy LLM. Each response is scored by the reward function $r_{\phi}(q, y)$ , producing a corresponding set of rewards $\mathbf{r} = \{r_1, r_2, \ldots, r_G\}$. Following \citet{shao2024deepseekmathpushinglimitsmathematical}, these rewards are then normalized by subtracting the group mean and dividing by the group standard deviation. The global advantages $A_{i,t}^\tau$ for all tokens within the same output are then set to this normalized value:
\begin{equation}
A_{i,t}^\tau = \hat{r}_i = \frac{r_i - \text{mean}(\mathbf{r})}{\text{std}(\mathbf{r})}.
\label{Eq:global-advantage}
\end{equation}
This design ensures that each trajectory shares a global and base advantage signal across all turns.


\subsection{Retrospective Critic: Action Advantage Estimate}
\begin{figure}[htbp]
  \centering
  \includegraphics[width=0.85\linewidth]{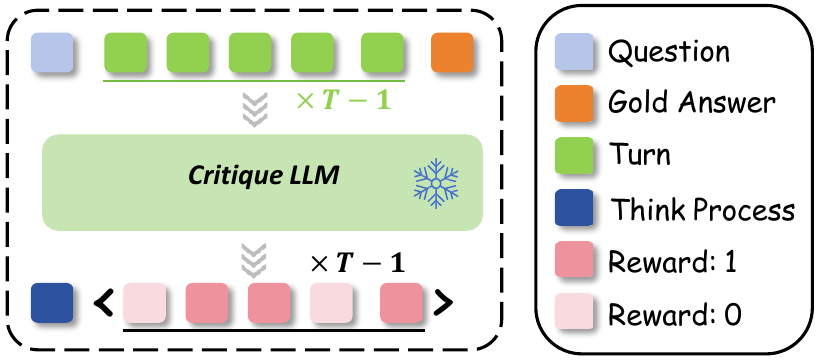} 
\caption{\textbf{Overview of the Critique  LLM.} 
The entire reasoning trajectory, along with the gold answer, is fed into the critique LLM, 
which evaluates each action as either \texttt{Good} or \texttt{Bad} and generates a corresponding turn-level reward sequence after thinking.}
  \label{fig:reward_model}
\end{figure}
We next describe the design of the turn-level reward model, which provides dense feedback during multi-turn reasoning.
Given a question $x$, the policy LLM $\pi_\theta$ samples a multi-turn trajectory
\[
y_i=\{(a_{i,t},\,c_{i,t})\}_{t=1}^{T},
\]
where $a_{i,t}$ is the action at the turn $t$ and $c_{i,t}$ is the information returned by the search engines
in response to $a_{i,t}$. During training, we assume access to privileged information: the ground-truth answer $o_{\text{gold}}$ and the full trajectory.
A frozen critique LLM $C_\phi$ takes $(x, y_i, o_{\text{gold}})$ as input and returns a per-turn judgment $\{\ell_{i,t}\}^T_{t=1}, \ell_{i,t}\in\{\texttt{Good},\texttt{Bad}\}$ after thinking (Fig.~\ref{fig:reward_model}). In this process, the critique LLM adopts a retrospective (hindsight) perspective to evaluate whether each search action has guided the reasoning toward useful information or, conversely, produced misleading or redundant evidence. Based on this reflection, it judges the quality of each search action.

We map these judgment to turn-level rewards
\begin{equation}
r^a_{i,t}=
\begin{cases}
1, & \ell_{i,t}=\texttt{Good},\\
0, & \text{otherwise},
\end{cases}
\label{Eq:turn-reward}
\end{equation}
This yields a reward sequence $\mathbf{r}^a_i=\{r^a_{i,t}\}_{t=1}^{T_i}$. A complete scoring example can be found in Fig.~\ref{fig:reard_case}. Details of the prompt for the critique LLM $C_\phi$ can be found in Appendix~\ref{reward_prompt}. 

To allocate credit across turns, we normalize per-turn rewards into a turn-level advantage
\begin{equation}
{A}^a_{i,t}
=\frac{r^a_{i,t}}{\sum_{u=1}^{T_i} r^a_{i,u}+\varepsilon},\qquad \varepsilon>0,
\label{Eq:turn-advantage}
\end{equation}
which avoids numerical issues when all turns are judged \texttt{Bad}.

This approach provides the model with a form of hindsight supervision, enabling the critique model to infer the contribution of each action using the asymmetric and privileged information that would not be available during forward decision-making.

\subsection{Policy Optimization}
Intuitively, $\tilde{A}^a_{i,t}$ provides a dense, hindsight credit distribution using privileged information, while $A_{i,t}^\tau$ preserves the global task signal. Therefore, we adopt GRPO as the backbone algorithm and replace its original advantage function with a new hybrid advantage during training:  
\begin{equation}
\widetilde{A}_{i,t} = \alpha A_{i,t}^a + (1 - \alpha) A_{i,t}^\tau,
\label{Eq:advantage}
\end{equation}
where $\alpha$ is a weighting coefficient that balances the global advantage $A_{i,t}^\tau$ and the turn-level advantage $A_{i,t}^a$.
{\small
\begin{equation}
\begin{aligned}
\mathcal{J}_{}(\theta) = \mathbb{E}_{q,\{y_i\} \sim \pi_{\text{old}}} \Bigg[ 
\frac{1}{G} \sum_{i=1}^G\sum_{t=1}^{|y_i|} \widetilde{\mathcal{L}}_{i,t} 
- \beta \, \mathbb{D}_{\text{KL}}(\pi_\theta \| \pi_{\text{ref}}) 
\Bigg]
\end{aligned}
\label{Eq:criticsearch_loss}
\end{equation}}
where the loss $\widetilde{\mathcal{L}}_{i,t}$ is defined as:
\begin{equation}
\small
\widetilde{\mathcal{L}}_{i,t} = \min \left( w_{i,t} \textcolor{blue}{\widetilde{A}_{i,t}},\ \operatorname{clip}(w_{i,t},\ 1 - \epsilon,\ 1 + \epsilon) \textcolor{blue}{\widetilde{A}_{i,t}} \right)
\label{Eq:loss}
\end{equation}


\section{Experiment}

\begin{table*}[!t]
\centering
\scriptsize
\renewcommand{\arraystretch}{1.2} 
\setlength{\tabcolsep}{4.8pt} 
\resizebox{0.8\textwidth}{!}{%
\begin{tabular}{lcccccccc|cc}

\toprule
 \multirow{2}{*}{\textbf{Method}}  & \multicolumn{2}{c}{\textbf{HotpotQA}$^{\circ}$} & \multicolumn{2}{c}{\textbf{2Wiki}$^{\diamond}$} & \multicolumn{2}{c}{\textbf{MuSiQue}$^{\diamond}$} & \multicolumn{2}{c}{\textbf{Bamboogle}$^{\diamond}$}& \multicolumn{2}{c}{\textbf{Average}} \\
\cmidrule(lr){2-3}\cmidrule(lr){4-5}\cmidrule(lr){6-7}\cmidrule(lr){8-9}\cmidrule(lr){10-11}
& \textit{EM} &\textit{F1} & \textit{EM} &\textit{F1}&\textit{EM}&\textit{F1}&\textit{EM}&\textit{F1}&\textit{EM}&\textit{F1} \\
\hline
\multicolumn{11}{l}{\textbf{\textit{Qwen2.5-3B-Base/Instruct}
}}\\
 Naive Generation & 14.5 & 23.7 & 24.9 & 35.6 & 1.8 & 7.9 & 3.0 & 8.6 & 11.1 & 19.0 \\
 Search-o1 & 24.0 & 32.6 & 20.7 & 30.9 & 4.5 & 11.7 & 31.6 & 43.6 & 20.2 & 29.7 \\
 Search-R1-base & 27.2 & 36.1 & 24.8 & 29.6 & 8.1 & 14.6 & 17.6 & 27.0 & 19.4 & 24.9 \\
 Search-R1-instruct & 30.4 & 40.1 & 29.3 & 35.2 & 12.0 & 18.8 & 24.0 & 34.4 & 23.9 & 32.1 \\
 ZeroSearch-base & 26.0 & 35.4 & 23.4 & 28.1 & 5.6 & 11.6 & 9.6 & 19.3 & 16.1 & 23.6 \\
 ZeroSearch-instruct & 26.5 & 35.5 & 23.3 & 27.8 & 5.9 & 12.1 & 14.4 & 24.3 & 17.5 & 25.0 \\
 StepSearch-base & 32.9 & 43.4 & \uline{33.9} & \uline{39.5} & \textbf{18.1} & \textbf{27.3} & 32.8 & 41.9 & 29.4 & 38.0 \\
 StepSearch-instruct & \uline{34.5} & \uline{45.2} & 32.0 & 38.5 & 17.4 & 26.1 & \uline{34.4} & \uline{45.2} & \uline{29.6} & \uline{38.3} \\

  \rowcolor{lightpurple} CriticSearch &\textbf{41.4}&\textbf{53.3}&\textbf{40.9}&\textbf{48.1}&\uline{18.0}&\uline{26.8}&\textbf{36.8} &\textbf{47.0} &\textbf{34.3} &\textbf{43.8}  \\
 
\bottomrule
 \multicolumn{11}{l}{\textbf{\textit{Qwen2.5-7B-Base/Instruct}}}\\
 Naive Generation & 18.7 & 29.1 & 24.6 & 35.2 & 2.7 & 8.3 & 12.3 & 24.2 & 14.6 & 24.2 \\
 Search-o1 & 19.3 & 28.8 & 18.1 & 28.9 & 5.3 & 12.7 & 30.2 & 42.7 & 18.2 & 28.3 \\
 Search-R1-base & \uline{43.2} & \uline{54.7} & 35.0 & 41.1 & 20.6 & 29.0 & {43.0} & 54.5 & 35.5 & 44.8 \\
 Search-R1-instruct & 39.4 & 50.2 & 31.2 & 37.6 & 18.1 & 26.2 & 38.4 & 50.1 & 31.8 & 41.0 \\
 ZeroSearch-base & 29.4 & 39.4 & 27.5 & 32.4 & 10.2 & 17.5 & 25.8 & 37.3 & 23.2 & 31.7 \\
 ZeroSearch-instruct & 32.5 & 43.2 & 30.9 & 37.0 & 12.0 & 20.4 & 26.7 & 40.9 & 25.5 & 35.4 \\
 ReasonRAG & 38.4 & 48.9 & \textbf{43.6} & \textbf{50.4} & 12.8 & 20.6 & 36.0 & 45.5 & 32.7 & 41.3   \\
 StepSearch-base & 38.0 & 49.3 & {38.5} & {45.0} & \uline{21.6} & \textbf{32.4} & \uline{46.7} & \uline{57.3} & \uline{36.2} & \uline{46.0} \\
 StepSearch-instruct & 38.6 & 50.2 & 36.6 & 43.1 & \textbf{22.6} & \uline{31.2} & 40.0 & 53.4 & 34.5 & 44.5 \\
   \rowcolor{lightpurple} CriticSearch & \textbf{44.2} & \textbf{56.0} & \uline{42.8} & \uline{50.1} & 19.4 & 28.1 & \textbf{47.2} & \textbf{59.2} & \textbf{38.4} & \textbf{48.4} \\


\bottomrule
\end{tabular}
}
\caption{\textbf{The main results of CriticSearch.} \textbf{Bold} indicates the top-performing result, while
\uline{underline} denotes the second-best. $^{\circ}$/$\diamond$ represents in-domain/out-of-domain benchmarks. \textit{We employ critique models with the same parameter scale as the policy models \,(i.e. Qwen2.5-3B-Instruct for Qwen2.5-3B-Base and Qwen2.5-7B-Instruct for Qwen2.5-7B-Base) in CriticSearch training.}  CriticSearch versus other algorithms on the same question are provided in Appendix~\ref{app:case_study}.}
\label{tab:main-results}
\end{table*}

\subsection{Setup}

\paragraph{Training}
We implement our method based on the Search-R1\citep{jin2025searchr1trainingllmsreason} framework.
We conduct experiments on two models from the Qwen-2.5 series \citep{qwen2025qwen25technicalreport}. The policy models are Qwen-2.5-3B-Base and Qwen-2.5-7B-Base, while the critique models are Qwen-2.5-3B-Instruct and Qwen-2.5-7B-Instruct, respectively. The training dataset is HotpotQA \citep{yang2018hotpotqadatasetdiverseexplainable}.

To maintain alignment with prior work (\citealp{jin2025searchr1trainingllmsreason}; \citealp{wang2025stepsearchignitingllmssearch}), we use the 2018 Wikipedia dump \citep{karpukhin2020densepassageretrievalopendomain} as the corpus and employ E5 \citep{wang2024textembeddingsweaklysupervisedcontrastive} as the retriever during training.
At each retrieval step, we uniformly sample $k=3$ documents.
The maximum context length is set to 4K tokens, with a maximum of four turns of searching. The same configuration is adopted during evaluation. More details on experimental setups can be found in Appendix~\ref{app:exp-setups}.

\paragraph{Evaluation}
We evaluate our search agent primarily on four multi-hop Q\&A datasets, including 
HotpotQA \citep{yang2018hotpotqadatasetdiverseexplainable}, 
2WikiMultiHopQA \citep{ho2020constructingmultihopqadataset}, 
MuSiQue \citep{trivedi2022musiquemultihopquestionssinglehop}, and 
Bamboogle \citep{press2023measuringnarrowingcompositionalitygap}.
We adopt the canonical Exact Match (EM) and word-level F1 scores as two evaluation metrics to ensure fair comparison.

\paragraph{Baselines}
To effectively evaluate the efficacy of CriticSearch, we compare it against comprehensive baselines:
(1)\textbf{Direct Inference:} Naive generation \citep{wei2023chainofthoughtpromptingelicitsreasoning};
(2)\textbf{Inference with Retrieval:} 
Search-o1 \citep{li2025searcho1agenticsearchenhancedlarge};
(3)\textbf{Agentic RL:} Existing outstanding reinforcement learning methods combined with external search engines, including Search-R1 \citep{jin2025searchr1trainingllmsreason}, ZeroSearch \citep{sun2025zerosearchincentivizesearchcapability}, 
ReasonRAG \citep{zhang2025processvsoutcomereward},
and StepSearch \citep{wang2025stepsearchignitingllmssearch}.

\subsection{Main Results}
\textbf{CriticSearch consistently outperforms all baseline methods.} 
The main comparative results between CriticSearch and the baseline methods across the four datasets are summarized in Tab.~\ref{tab:main-results}. 
CriticSearch consistently outperforms all baselines on both in-domain multi-hop benchmarks (HotpotQA) and out-of-domain benchmarks (MuSiQue, 2WikiMultiHopQA, and Bamboogle), demonstrating strong robustness and generalization capability. Comparative case studies of CriticSearch versus other algorithms on the same question are provided in Appendix~\ref{app:case_study}.
\begin{figure}[!t]
  \centering
  \includegraphics[width=1\linewidth]{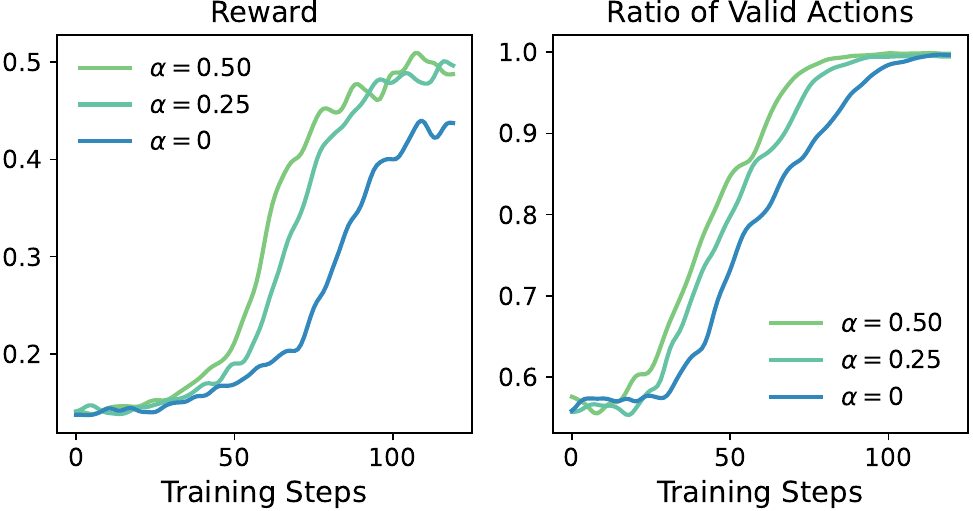} 
  \caption{\textbf{Ablation study on the weight of the turn-level advantage $\alpha$.} 
A larger $\alpha$ leads to faster convergence during the training phase.}
  \label{fig:coverage}
\end{figure}

\begin{figure*}
    \centering
    \includegraphics[width=0.85\linewidth]{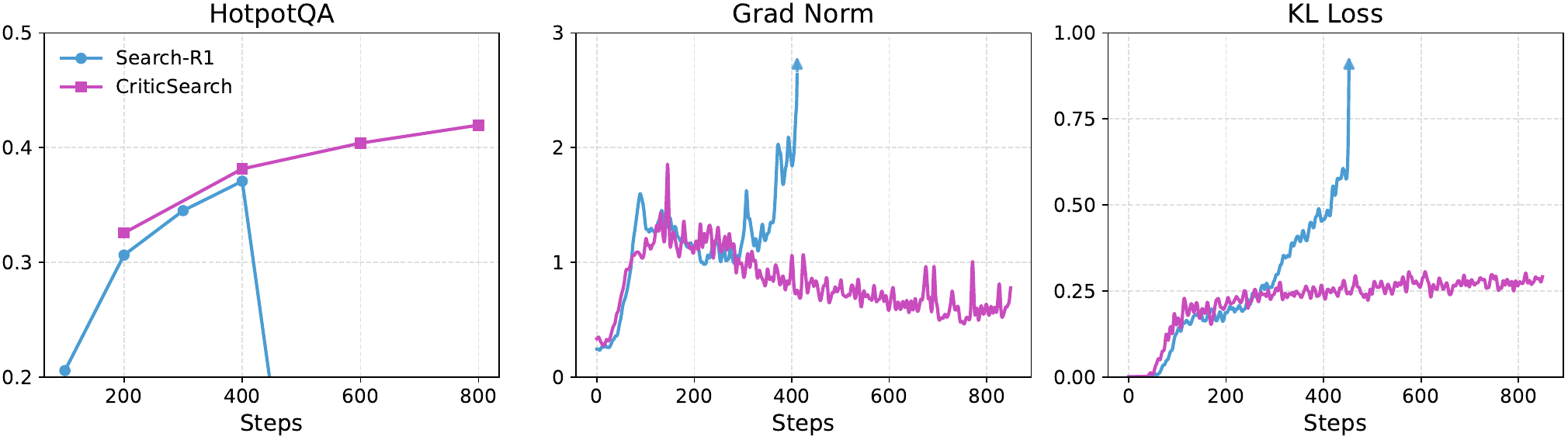}
    \caption{\textbf{Training stability and performance comparison.}
CriticSearch effectively constrains both the gradient norm and the KL loss within a stable range, preventing divergence and ensuring smooth optimization (right). 
This improved stability enables CriticSearch to sustain longer training and achieve higher accuracy.}
    \label{fig:collapse}
\end{figure*}

\textbf{CriticSearch surpasses methods trained with sparse outcome rewards.} 
Compared to Search-R1 and ZeroSearch, which rely solely on sparse reward signals during training, CriticSearch achieves substantially better performance, demonstrating that incorporating dense, fine-grained rewards effectively improves training efficiency.

\textbf{CriticSearch surpasses existing dense-reward methods.} 
While existing dense-reward baselines occasionally outperform CriticSearch on specific datasets (e.g., StepSearch on MuSiQue, ReasonRAG on 2Wiki), their overall performance still lags behind that of CriticSearch.

\section{Further Analysis}
\subsection{Accelerate the Convergence of Training}
As shown in Fig.~\ref{fig:coverage}, we illustrate the reward curve and the ratio of valid actions in the early training stage. 
We observe that increasing the weight of the turn-level advantage term in Eq.~\ref{Eq:advantage} leads to faster convergence in the training stages. 
This phenomenon is intuitive, as even failed reasoning trajectories may contain beneficial actions, whereas successful ones can still include suboptimal actions. 
By incorporating the turn-level advantage, the model can more precisely assign learning signals to individual search actions within each trajectory, thereby enhancing the LLM agent’s search capability and training efficiency, and ultimately accelerating overall convergence. 

\begin{table*}[h!t]
\centering
\resizebox{0.8\textwidth}{!}
{%
\begin{tabular}{cccccc}

\toprule

{\textbf{Method}} &  {\textbf{Critique model}} & {\textbf{HotpotQA}} & {\textbf{2Wiki}} & {\textbf{MuSiQue}} & {\textbf{Bamboogle}} \\

\hline
 \multicolumn{6}{l}{\textbf{\textit{Qwen2.5-3B-Base}
}}\\
 CriticSearch & \textit{None} & 38.1 & 39.0 & 14.1 & 32.8 \\
 CriticSearch &\textit{Qwen2.5-3B-Instruct}&{41.4}&{40.9}&{18.0}&{36.8}\\
 \rowcolor{lightpurple} CriticSearch &\textit{Qwen3-30B-A3B-Instruct-2507}& \textbf{41.9}&\textbf{42.4}&\textbf{18.6}&\textbf{43.2}\\

\bottomrule
 \multicolumn{6}{l}{\textbf{\textit{Qwen2.5-7B-Base}
}}\\
 CriticSearch & \textit{None} &41.8&40.0&18.7&40.8 \\
 CriticSearch &\textit{Qwen2.5-7B-Instruct}& 44.2 & 42.8 & 19.4 & \textbf{47.2} \\ 
 \rowcolor{lightpurple} CriticSearch &\textit{Qwen3-30B-A3B-Instruct-2507}& \textbf{46.1} & \textbf{43.2} & \textbf{20.5} & 40.8  \\ 

\bottomrule
\end{tabular}
}
\caption{\textbf{Ablation study on the size of critique LLM.} 
 A larger critique LLM yields  better performance.}
\label{tab:ablation-reward-model}
\end{table*}

\subsection{Mitigating Premature Training Collapse}

As shown in Fig.~\ref{fig:collapse}, we illustrate  the evaluation scores, gradients, and KL divergence during training. 
Under sparse rewards, the algorithm collapses after around 400 steps, whereas incorporating the turn-level advantage effectively mitigates this collapse and suppresses abnormal fluctuations in both the KL loss and gradients, allowing performance to continue improving.

When training with only sparse outcome rewards, each action within a correct trajectory receives a reward of 1, while all actions in incorrect trajectories receive 0. 
From the perspective of turn-level rewards, this introduces a highly noisy turn-level reward model that may misclassify poor actions as high-reward or penalize good actions with low scores. 
Such noise drives the LLM to optimize toward unreasonable search paths, producing redundant or inefficient behaviors misaligned with human preferences. 
The optimization over noisy reward signals during training causes the model to deviate from the human-aligned distribution represented by the reference model (e.g., Qwen-2.5-3B-base) \citep{liu2024improvingmultistepreasoningabilities,wang2025improving}. 
Consequently, when suboptimal actions receive undeservedly high rewards, the model distribution diverges from that of the reference model, leading to a continuously increasing KL loss and, ultimately, training collapse. 
In contrast, turn-level rewards provide more accurate supervision for each search action.  
By maintaining stable and informative reward signals throughout training, they effectively alleviate the collapse phenomenon.

\subsection{Ablation study on the Size of Critique }
Next, we conduct an ablation to verify whether a larger critique LLM produces higher-quality dense rewards, thereby training the policy LLM more effectively (Tab.~\ref{tab:ablation-reward-model}). We compare three settings for supervising the same policy model (Qwen2.5-3/7B-Base): 1) a small critique model (Qwen2.5-3/7B-Instruct), 2) a larger critique model (Qwen3-30B-A3B-Instruct-2507 \citep{yang2025qwen3technicalreport}), and 3) no critique model (i.e., no turn-level reward). As shown in Tab.~\ref{tab:ablation-reward-model}, the policy trained with the larger critique model achieves higher average performance across four multi-hop Q\&A datasets. These results provide evidence that a larger critique model provides more reliable dense feedback and more effectively assists policy learning, and validating the usefulness of the critique model. A detailed analysis of the accuracy of the trun-level reward is provided in Section \ref{sec:accuracy}.

\subsection{Accuracy analysis of Critique Model}
\label{sec:accuracy}

\begin{table*}[h!t]
\centering

\resizebox{0.65\textwidth}{!}
{%
\begin{tabular}{cccccc}

\toprule
{\textbf{Method}} & {\textbf{HotpotQA}} & {\textbf{2Wiki}} & {\textbf{MuSiQue}} & {\textbf{Bamboogle}}& {\textbf{Average}} \\
\hline
 \multicolumn{5}{l}{\textbf{\textit{Qwen2.5-7B-Base}
}}\\
 CriticSearch-Sparse  &41.8&40.0&18.7&40.8& 35.3\\
 CriticSearch-NG  & \textbf{44.9} & \uline{40.7} & \uline{19.2} & \uline{41.6}& \uline{36.6}  \\ 
 \rowcolor{lightpurple}CriticSearch & \uline{44.2} & \textbf{42.8} & \textbf{19.4}& \textbf{47.2} & \textbf{38.4}\\ 

\bottomrule
\end{tabular}
}
\caption{\textbf{Ablation study on the privileged information of the critique LLM.} CriticSearch-NG is the variant where the gold answer is removed from the critique LLM input during training. CriticSearch-Sparse is the variant trained solely with the outcome reward.}
\label{tab:ablation-gold_answer}
\end{table*}

We sampled 20 trajectories and, for each turn within a trajectory, assigned scores using 1) different critique models, including Gemini-2.5-pro \citep{comanici2025gemini25pushingfrontier}, Qwen3-30B-A3B-It, GPT-4o \citep{openai2024gpt4ocard}, and Qwen2.5-7B-It; 2) Monte Carlo estimates; 3) Outcome Reward (i.e., using the outcome reward as the turn-level reward), and 4) human annotations.   We then analyzed the agreement among these scoring methods.
Assuming human annotations as the ground truth, we observe that larger and more capable critique models exhibit higher alignment with human judgments.    Specifically, Gemini-2.5-Pro achieves the highest agreement with human annotations, reaching approximately 80\% similarity.
Qwen2.5-7B-It achieves a level of agreement with human scores comparable to Monte Carlo estimation;  however, the Monte Carlo approach requires many more rollouts to obtain reasonably accurate estimates and is therefore more resource-intensive.   Notably, the larger Qwen3-30B-A3B-It model surpasses the Monte Carlo approach in its alignment with human annotations. 
In contrast, using the outcome reward as the turn-level score results in the lowest agreement with human judgments. Further details on this experiment are provided in Appendix~\ref{app:exp-similarity}.
\begin{figure}
    \centering
    \includegraphics[width=1\linewidth]{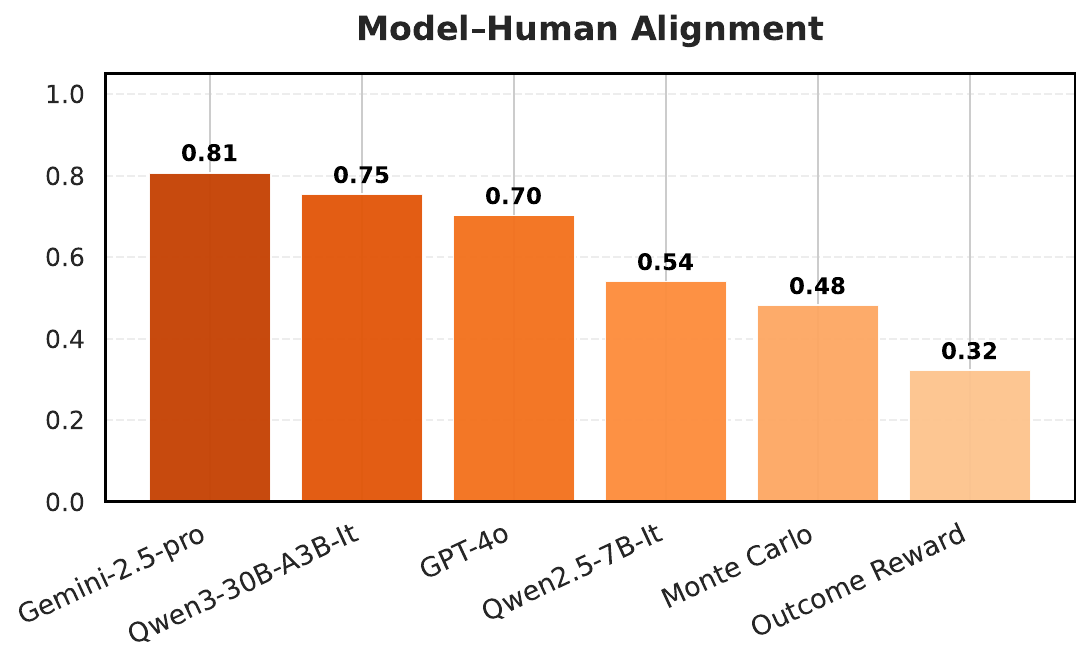}
    \caption{\textbf{Alignment Across Dense-Reward Methods.}
Higher similarity to human annotations indicates more accurate turn-level rewards.}
    \label{fig:similarity}
\end{figure}

\subsection{Ablation study on the privileged information of Critique LLM}
\begin{figure}[h]
    \centering
    \includegraphics[width=1\linewidth]{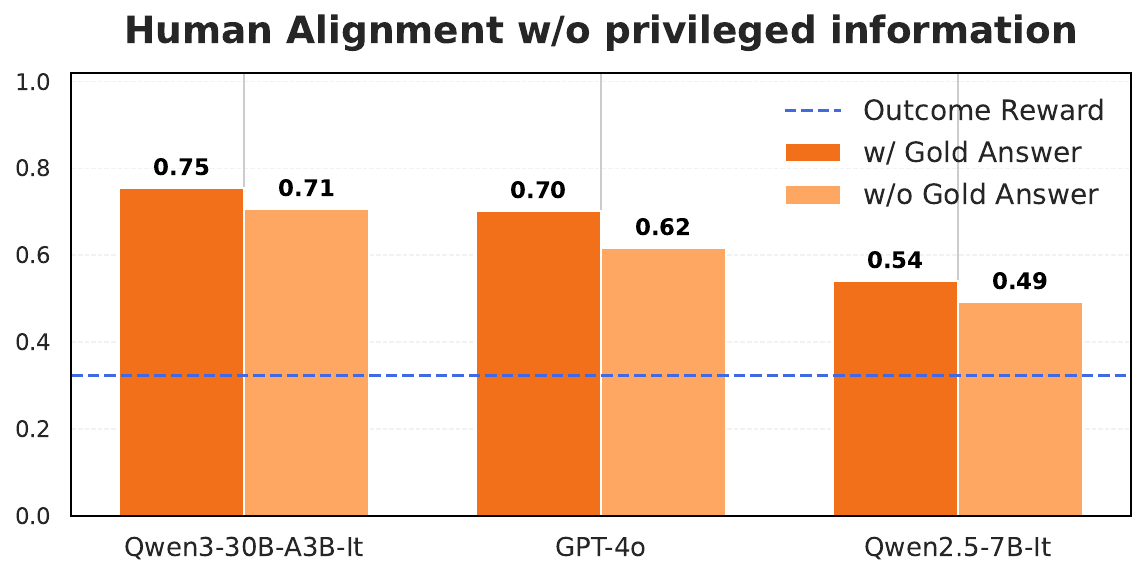}
    \caption{Removing the gold answer from the critique LLM slightly reduces its alignment with human annotations, indicating a moderate decline in reward accuracy. The blue dashed line indicates the alignment when using the outcome reward as the turn-level reward.}
    \label{fig:wo_gold}
\end{figure}

To evaluate the effect of privileged information in the critique LLM (i.e., the gold answer and future trajectory), we ablated the gold answer from its input and measured the accuracy of the resulting turn-level rewards. 
As shown in Fig.~\ref{fig:wo_gold}, excluding the gold answer leads to a moderate decline in reward accuracy. 
Nevertheless, with the benefit of a retrospective assessment, the critique LLM can still capture the semantics of the search actions. Therefore, its turn-level rewards remain more accurate than outcome-based turn-level rewards.
Consistently, the ablation results in Table~\ref{tab:ablation-gold_answer} indicate that removing the gold answer causes some performance degradation, yet the model still outperforms the variant without a critique LLM (i.e., using only the outcome reward).

\section{Conclusion}
This paper introduces CriticSearch, a fine-grained credit assignment method that leverages large language models to assist policy training.   By introducing a retrospective critic mechanism, we employ a lightweight turn-level binary evaluator that delivers more effective dense feedback while mitigating training instability.  Extensive experiments and analyses demonstrate that CriticSearch consistently improves learning efficiency and stability, highlighting its effectiveness and robustness across diverse multi-hop Q\&A benchmarks.

\section{Limitations}
The inclusion of the Critique model increases the memory footprint required for model loading, while the additional reward computation introduces a small amount of extra training overhead compared to GRPO (Appendix~\ref{app:train_cost}). 
Moreover, the proposed method has so far been validated only on search-related tasks with a limited number of reasoning turns. 
Future work may explore reusing the policy model as the critique model during training and further validate the method’s effectiveness in broader agentic RL scenarios.

\bibliography{acl_latex}

\clearpage
\appendix

\section{Related Works}
\subsection{Agentic RL with Search Engines}
Recently, Reinforcement Learning has emerged as a promising framework for enhancing the reasoning capabilities of large language models, demonstrating strong proficiency in complex reasoning tasks \citep{shao2024deepseekmathpushinglimitsmathematical,deepseekai2025deepseekr1incentivizingreasoningcapability}. 
To address the limited access of LLMs to up-to-date external knowledge, several studies \citep{jin2025searchr1trainingllmsreason, wang2025stepsearchignitingllmssearch, chen2025researchlearningreasonsearch, zhang2025nemotronresearchtooln1exploringtoolusinglanguage} have explored RL-based approaches that integrate external search engines to improve information retrieval effectiveness. 
Beyond this, a number of works aim to further enhance search-agent performance from multiple perspectives \citep{gao2025turnsunlockinglonghorizonagentic, sun2025simpledeepsearcherdeepinformationseeking}. 
Some methods refine underlying mechanisms by summarizing previous interactions to shorten the effective context length \citep{zhou2025mem1learningsynergizememory}, introducing fine-grained reward functions for more efficient optimization \citep{wang2025stepsearchignitingllmssearch, wei2025reinforcingmultiturnreasoningllm,zhang2025letslearningthinkandsearchprocessandoutcome,qian2025toolrlrewardtoollearning}, and leveraging multi-agent collaboration for improved coordination and reasoning \citep{li2025chainofagentsendtoendagentfoundation, dong2025toolstarempoweringllmbrainedmultitool}. 
Other studies focus on optimizing interaction strategies and data utilization. 
For example, DeepResearch integrates search agents with real-world search environments \citep{zheng2025deepresearcherscalingdeepresearch}, while ZeroSearch replaces traditional search engines with LLM-based simulated retrieval, and several works \citep{sun2025simpledeepsearcherdeepinformationseeking, gao2025turnsunlockinglonghorizonagentic} investigate improvements in multi-task settings and with richer data resources. 
Our work is similar to StepSearch, introducing a fine-grained reward mechanism for credit assignment under sparse rewards. 
Unlike StepSearch, which relies on curated ground-truth references for supervision, CriticSearch adopts a streamlined, general framework that leverages the existing strong reasoning ability of the LLM to evaluate intermediate actions and assign credit.

\subsection{Reward Models in LLM Reinforcement Learning}
The Reinforcement Learning with Verifiable Rewards (RLVR) paradigm \citep{lambert2025tulu3pushingfrontiers} has greatly advanced large language models’ reasoning in mathematics and code generation, as shown by recent systems such as \citet{deepseekai2025deepseekr1incentivizingreasoningcapability}, \citet{kimiteam2025kimik2openagentic}, and \citet{5team2025glm45agenticreasoningcoding}.
However, outcome-based rewards remain inherently sparse, offering supervision only at the end of reasoning trajectories. This limitation can reward spurious reasoning paths that coincidentally yield correct answers and cause severe credit-assignment issues \citep{lightman2023letsverifystepstep}, especially in multi-turn agentic RL where models must coordinate sequential tool calls.
To address these problems, prior research has explored complementary directions:
1) designing explicit process-level rewards that score intermediate reasoning or tool usage \citep{qian2025toolrlrewardtoollearning, wang2025stepsearchignitingllmssearch}. However, such approaches are often tailored to specific tools or datasets, resulting in limited generalizability.
2) employing outcome-driven reward inference that propagates final rewards backward through trajectory grouping or branch analysis \citep{feng2025groupingrouppolicyoptimizationllm,dong2025agenticreinforcedpolicyoptimization}. These approaches often suffer from high variance and require large amounts of rollout for stable estimation. 
3) training turn-level critics that provide local supervision for each tool-calling step \citep{zhou2025sweetrltrainingmultiturnllm,wang2025sparlreinforcingllmagents}.
Similar to the third approach, our method introduces a lightweight turn-level binary evaluator that simply labels each tool invocation as either good or bad, thereby reducing the risk of reward hacking. Unlike training-based critics, the evaluator remains frozen during learning and requires no additional training. This not only avoids the instability and computational overhead associated with joint critic optimization but also eliminates the need for extra training data, thereby enhancing both efficiency and generalizability. This design effectively bridges the gap between sparse outcome supervision and dense step-level feedback, yielding a more stable and scalable online RL framework for agentic reasoning.

\begin{figure*}
    \centering
    \includegraphics[width=1\linewidth]{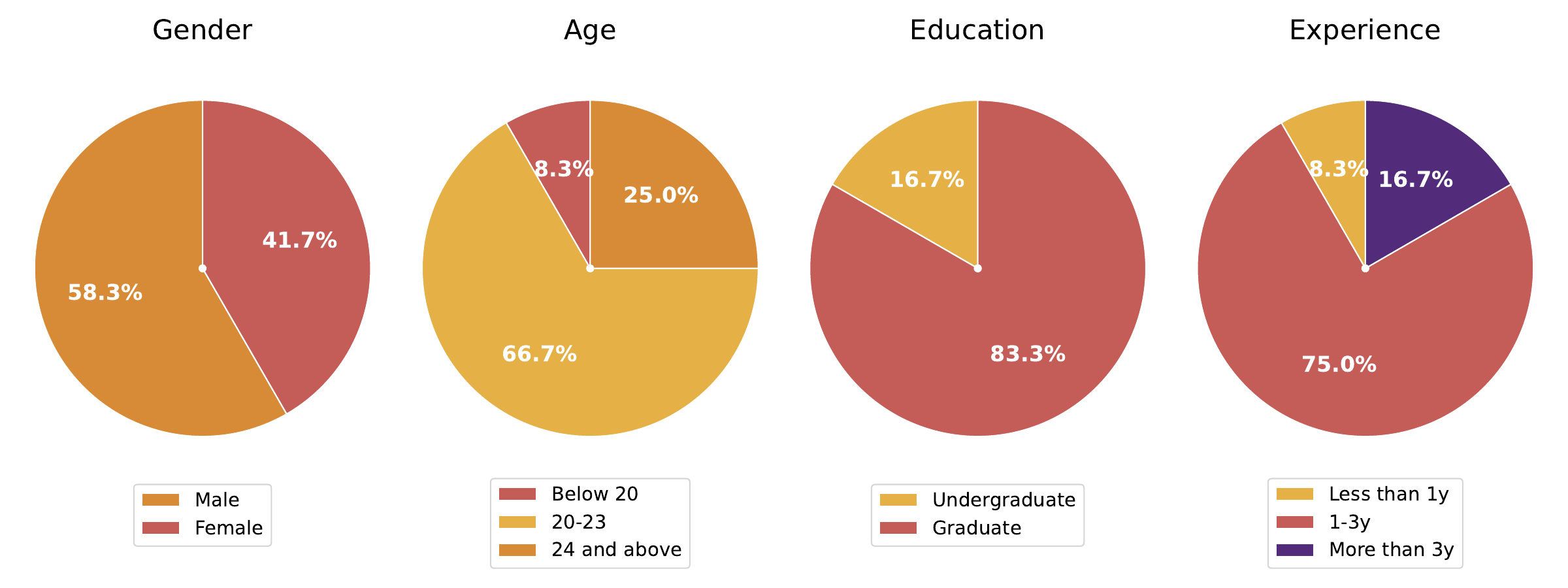}
    \caption{\textbf{Distribution of participants involved in the human evaluation experiment is presented.} Each participant was asked to provide basic information, including gender, age, education level, and experience of AI. We present the distribution of these information categories in the form of pie charts.}
    \label{fig:human-distribution}
\end{figure*}

\section{Experiment Setups}\label{app:exp-setups}
Our training is conducted on the VerL\citep{Sheng_2025} framework. For GRPO training, we set the group size $G$ to 5 and the policy LLM learning rate $5\times 10^{-7}$ for all models. The rollout temperature is set to $1.0$ \,. We train for a total of 800 steps, with a warm-up ratio of $0.35$ \,. The total batch size, mini batch size, and micro batch size are set to 128, 64, and 32 respectively. The entropy regularization coefficient and the KL loss coefficient are both set to $1\text{e-}3$, and the PPO clip ratio $\epsilon$ to $0.2$. 

We set $\lambda_f$ as $0.2$ and $\lambda_r$ as $0.1$ for all experiments. We sweep the weighting coefficient $\alpha \in \{0.25, 0.5, 0.75\}$ and observe that the optimal setting is consistently $\alpha=0.25$ for all models. 
All training jobs are conducted on a node with 8 NVIDIA H100 GPUs. Early stopping is applied when training collapse is detected from the reward curve. For the results in Tab.~\ref{tab:main-results} and Tab.~\ref{tab:ablation-reward-model}, we report either the final checkpoint at step $800$ or the latest checkpoint before collapse.

To ensure a fair comparison, we adopt the original open-source model checkpoints and their published prompt configurations for all baselines.
For consistency, they are evaluated using the same retrieval configuration as our agent.

\section{Additional Results}

\subsection{Training Cost}
\label{app:train_cost}

\begin{table}[h!t]
\centering
\resizebox{0.47\textwidth}{!}
{%
\begin{tabular}{ccc}

\toprule

{\textbf{Method}} &  {\textbf{Critique model}} & {\textbf{Time}}  \\

\hline
 \multicolumn{3}{l}{\textbf{\textit{Qwen2.5-3B-Base}
}}\\
 Search-R1 & \textit{-} & 68.4  \\
 CriticSearch &\textit{Qwen2.5-3B-Instruct}&79.5\\
 CriticSearch &\textit{Qwen3-30B-A3B-Instruct-2507}& 88.9\\

\bottomrule
 \multicolumn{3}{l}{\textbf{\textit{Qwen2.5-7B-Base}
}}\\
  Search-R1 & \textit{-} &114.7 \\
 CriticSearch &\textit{Qwen2.5-7B-Instruct}& 110.9 \\ 
 CriticSearch &\textit{Qwen3-30B-A3B-Instruct-2507}& 140.8  \\ 

\bottomrule
\end{tabular}
}
\caption{\textbf{The training time.} We analyze the per-step training time (s) of different methods. When using a critique model of the same size as the policy model, CriticSearch introduces almost no additional training time compared with Search-R1.}
\label{tab:time}
\end{table}

\begin{figure*}[t]
\centering
  \includegraphics[width=\linewidth]{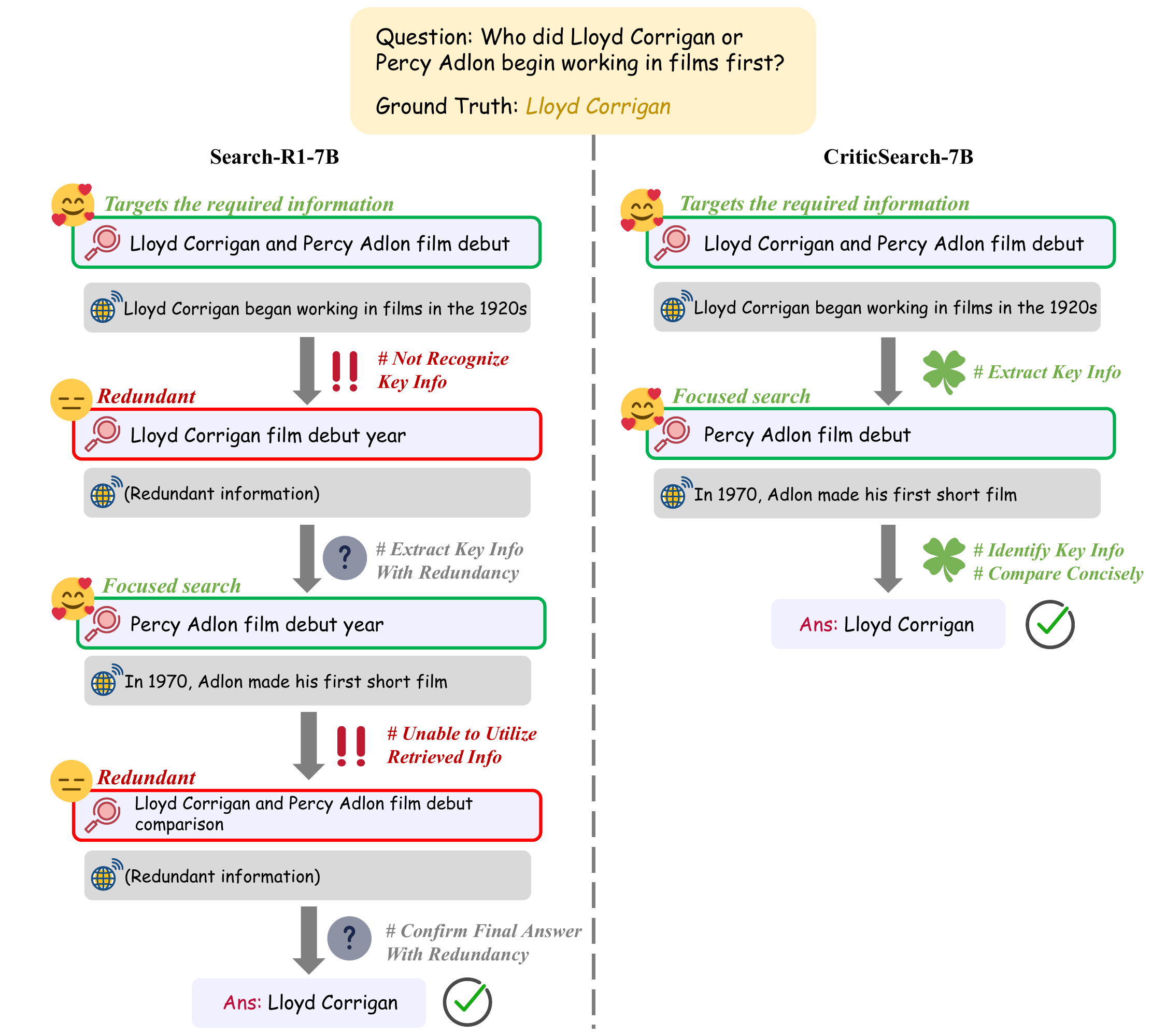}
    \caption{Side-by-side trajectories of Search-R1 (left, generated by \textit{SearchR1-qwen2.5-7b-grpo-v0.3}) and CriticSearch (right) on the same query. Each trajectory shows multi-turn interactions with the search engine (actions, responses, and final answer). Although both Search-R1 and CriticSearch answer the query correctly, our method (right) uses fewer, more focused queries and reaches a logically structured answer with minimal redundancy.}
  \label{app:case-study_4}
\end{figure*}

Tab.~\ref{tab:time} reports the average per-step training time of different methods. 
Across various search agents, we observe that although our approach introduces an additional retrospective critic mechanism compared with Search-R1, it does not significantly increase the per-step training time.
When using a critique model of the same size as the policy model, CriticSearch introduces almost no additional training time compared with Search-R1. Especially when using Qwen2.5-7B-Base as the policy model and Qwen2.5-7B-Instruct as the critique model, CriticSearch even achieves a shorter per-step time than Search-R1. We attribute this to the fact that Search-R1 lacks fine-grained rewards to penalize low-quality actions, leading to redundant and repetitive search action during rollouts  (Fig.~\ref{app:case-study_4}), which in turn increases training time. 

The reported training times were measured on a node equipped with 8 NVIDIA H100 GPUs, and we follow the same training configurations described in Appendix~\ref{app:exp-setups}.

\subsection{Accuracy analysis of Critique Model}
\label{app:exp-similarity}
We sampled 20 trajectories using the SearchR1-qwen2.5-7b-grpo-v0.3 model. For the Monte Carlo rearward estimates, we used the same model to roll out 10 trajectories forward from each action and took the average as the estimated action reward. For Gemini-2.5-Pro \citep{comanici2025gemini25pushingfrontier} and GPT-4o \citep{openai2024gpt4ocard}, we adopted the same prompt (Appendix~\ref{reward_prompt}) as Qwen2.5-7B-It \citep{yang2025qwen3technicalreport}  to perform reward generation.

For human-labeled rewards, we recruited 12 human evaluators to assess the quality of each action along the trajectories. Each evaluator was asked to provide basic demographic information to help us understand the evaluator distribution. The demographic fields included gender, age, education, and experience with AI. The distribution of evaluators for each task is shown in Fig.~\ref{fig:human-distribution}.

\section{Case Study}
\label{app:case_study}
We observe the sampled trajectories during the CriticSearch training process, and representative examples are presented in Tab.~\ref{app:case-study_1}, Tab.~\ref{app:case-study_2} and Tab.~\ref{app:case-study_3}.
All cases are obtained from the model trained on Qwen2.7-7B-Base.
In particular, we further compare the trajectory generated by the baseline Search-R1 and our CriticSearch model on the same question, as shown in Fig.~\ref{app:case-study_4}.
Each trajectory contains multi-turn interactions between the model and the external search engine.
Our CriticSearch method introduces retrospective critic fine-grained credit assignment at the process level, where each search action is judged by an external reasoning-capable LLM according to its usefulness to the final answer.
This dense supervision enables the policy to assign credit at the step level—encouraging informative searches while penalizing redundant or misleading ones.
As a result, in the later stage, the model naturally learns to issue concise, logically progressive queries (e.g., decomposing multi-entity comparisons or refining layered reasoning chains), forming more efficient trajectories with minimal redundancy and stable convergence.

\begin{table}[t]
\centering
\small
\resizebox{\columnwidth}{!}{
\begin{tabular}{@{}p{\columnwidth}@{}}
\toprule
\textbf{Question:} Aftermath was based on a real life midair plane collision that happened on what date?\\
\midrule
\textbf{Ground Truth:} ['1 July, 2002']\\
\midrule

\think{I need to find out when the real life midair plane collision that Aftermath was based on happened. }\\
\search{Aftermath midair plane collision date}\\
\info{\textcolor{gray}{Doc 1(Title: "Mid-air collision") The US Defense Safety Oversight Council established the official U.S. civil–military mid-air collision prevention website in 2006... Doc 2(Title: "Mid-air collision") The primary causes of mid-air collisions include miscommunication, navigational errors, and the absence of collision-avoidance systems... Doc 3(Title: "Mid-air collision") Several historical U.S. civil–military collision incidents led to the Federal Aviation Act of 1958 and later initiatives.}}\\

\think{I found information about midair collisions and their prevention, but I need to find the date of \textbf{the specific collision that Aftermath was based on}.}\\
\search{Aftermath \textbf{movie} midair collision date}\\
\info{Doc 1(Title: "Aftermath (2017 film)") \textcolor{red}{The film is based on the July 2002 Überlingen mid-air collision}... \textcolor{gray}{Doc 2(Title: "Aftermath (2017 film)") Aftermath (originally titled 478) is a 2017 American drama thriller film directed by Elliott Lester and written by Javier Gullón... Doc 3(Title: "Aftermath (2017 film)") In November 2015, Highland Film Group sold the film to different international distributors at the American Film Market.} } \\

\think{I found that Aftermath is based on \textbf{the July 2002 Überlingen mid-air collision}. Now I need to confirm the date of this collision.} \\
\search{Überlingen midair collision date} \\

\info{Doc1(Title: "2002 Überlingen mid-air collision") \textcolor{red}{On the night of 1 July 2002, Bashkirian Airlines Flight 2937}, ...  \textcolor{gray}{Doc 2(Title: "2002 Überlingen mid-air collision") It was being flown by two Bahrain-based pilots, 47-year-old British Captain... Doc 3(Title: "2002 Überlingen mid-air collision") The Tupolev aircraft continued descending despite TCAS advisories and that the accident prompted recommendations to upgrade TCAS systems...}  } \\

\think{I found the date of the Überlingen mid-air collision.} \\ 
 \answer{July 1, 2002} \\
\bottomrule
\end{tabular}
}
\caption{An example from the late stage of CriticSearch training. After obtaining invalid information in the first search, the model adjusts its query keywords and continues the subsequent search process. Red-font text denotes \textcolor{red}{valid information}; struck-through text indicates \sout{redundant content}; and gray text represents \textcolor{gray}{question-independent information}.}
\label{app:case-study_1}
\end{table}

\begin{table}[t]
\centering
\small
\resizebox{\columnwidth}{!}{
\begin{tabular}{@{}p{\columnwidth}@{}}
\toprule
\textbf{Question:} When was the college, for which Willie Fritz was head coach from 2014 to 2015, founded?\\
\midrule
\textbf{Ground Truth:} ['1906']\\
\midrule

\think{I need to determine when the college for which Willie Fritz was head coach from 2014 to 2015 was founded.}\\
\search{Willie Fritz head coach 2014 to 2015}\\
\info{ Doc 1(Title: "Willie Fritz") \textcolor{red}{From 2014 to 2015, he was head coach at Georgia Southern University}... \sout{Doc 2(Title: "Willie Fritz") From 2014 to 2015, he was head coach at Georgia Southern University...} \sout{Doc 3(Title: "Willie Fritz") ...On January 10, 2014, Fritz became the head football coach at Georgia Southern University}...
}\\

\think{I found information about Willie Fritz and his coaching positions, but I need to determine the founding year of the college he was head coach at from 2014 to 2015.}\\
\search{Georgia Southern University founded}\\
\info{Doc 1(Title: "Georgia Southern University") \textcolor{red}{Founded in 1906 as a land grant college}, Georgia Southern is part of the University System... 
\textcolor{gray}{Doc 2(Title: "Georgia State University") Georgia State University Georgia State University is a public research university in downtown Atlanta ...  Doc 3(Title: "Savannah State University") Savannah State University was founded as a result of the Second Morrill Land Grant Act of August 30, 1890.
}
}\\

\think{I found the founding year of Georgia Southern University. } \\ 
 \answer{1906} \\
\bottomrule
\end{tabular}
}
\caption{An example from the late stage of CriticSearch training. The model demonstrates the ability to perform progressive search actions in a natural manner. Red-font text denotes \textcolor{red}{valid information}; struck-through text indicates \sout{question-independent information}; and gray text represents \textcolor{gray}{redundant content}.}
\label{app:case-study_2}
\end{table}

\begin{table}[t]
\centering
\small
\resizebox{\columnwidth}{!}{
\begin{tabular}{@{}p{\columnwidth}@{}}
\toprule
\textbf{Question:} What rock band was formed in Oxford and reached 5 in the UK Charts?\\
\midrule
\textbf{Ground Truth:} ['Supergrass']\\
\midrule

\think{I need to determine which rock band was formed in Oxford and reached number 5 in the UK Charts. }\\
\search{rock band formed in oxford reached number 5 in the uk charts}\\
\info{ Doc 1(Title: "Oxford") \textcolor{red}{The most notable Oxford act is Radiohead, who all met at nearby Abingdon School, though other well known local bands include Supergrass...} \textcolor{gray}{ Doc 2(Title: "The Oxford Circle")The Oxford Circle was an American garage rock and psychedelic rock band from Davis... Doc 3(Title: "Oxford") Bands from over 30 years of the Oxford music scene's history feature in the documentary film ""Anyone Can Play Guitar?"".} 
}\\

\think{I found information about different bands from Oxford, but I need to find the specific band that reached number 5 in the UK Charts.}\\
\search{Radiohead reached number 5 in the uk charts}\\
\info{
Doc 1(Title: Radiohead) \textcolor{red}{""Street Spirit"" placed Radiohead in the top five for the first time}...
Doc 2(Title: Radiohead) \textcolor{red}{Radiohead are an English rock band from Abingdon, Oxfordshire, formed in 1985}... Doc 3(Title: "Radiohead discography") ""Amnesiac"", Radiohead's fifth studio album, was released in June 2001, \textcolor{red}{topping the UK charts} and producing successful singles including...
}\\

\think{I found that Radiohead is a rock band from Abingdon, Oxfordshire, not Oxford, but they reached number 5 in the UK Charts with the song "Street Spirit (Fade Out)".} \\ 
\search{Supergrass reached number 5 in the uk charts}\\
\info{
Doc 1(Title: "Supergrass")\textcolor{red}{The band's debut album, ""I Should Coco"" (May 1995), which entered the UK Albums Chart at number one. The album's fourth single, the double A-side release ""Alright""/""Time"", stayed in the UK Top Three for a month, peaking at number two} ...
Doc 2(Title: "Supergrass")\textcolor{gray}{Supergrass Supergrass were an English rock band, formed in 1993 in Oxford.} \sout{The band consisted of brothers Gaz}...
Doc 3(Title: "Supergrass")\textcolor{red}{The record entered the UK albums chart at number four and has since gone gold in the UK}.
}\\
\think{I found that Supergrass, a rock band from Oxford, reached number 5 in the UK Charts with the song "Alright". } \\
 \answer{Supergrass} \\
\bottomrule
\end{tabular}
}
\caption{An example from the late stage of CriticSearch training. The policy model demonstrates improved search reasoning and evidence localization ability. Red-font text denotes \textcolor{red}{valid information}; struck-through text indicates \sout{question-independent information}; and gray text represents \textcolor{gray}{redundant content}.}
\label{app:case-study_3}
\end{table}

\section{Algorithm Workflow}

In this section, we present a detailed description of the complete training workflow of \textit{CriticSearch} (Alg.~\ref{app:algo}).
\begin{algorithm}[t]
\caption{Training Stage of CriticSearch}
\normalsize
\begin{algorithmic}[1]
\REQUIRE Policy model $\pi_\theta$, Reference model $\pi_{\text{ref}}$, Frozen critic $C_\phi$, 
Training dataset $\mathcal{D}$, 
Hyperparameters $\lambda_f$, $\alpha$, $\beta$, $\epsilon$, $\varepsilon$
\ENSURE Optimized policy $\pi_\theta$

\vspace{1mm}
\STATE \textit{\textcolor{gray}{\# Rollout}}
\STATE Sample a batch $\mathcal{D}_b$ from $\mathcal{D}$
\FOR{each question $q \in \mathcal{D}_b$}
    \STATE Sample $G$ responses $\{y_i\}_{i=1}^G \sim \pi_\theta(q)$
    \STATE \
    \STATE \textit{\textcolor{gray}{\# Estimate Global Advantage}}
    \FOR{each response $y_i$}
        \STATE Compute outcome reward $r_i = r_\phi(q, y_i)$
    \ENDFOR
    \FOR{each response $y_i$}
        \STATE Computet global advantage $A_{i,t}^\tau$ via Eq.~\ref{Eq:global-advantage}
    \ENDFOR
    \STATE \
    \STATE \textit{\textcolor{gray}{\# Estimate Turn-level Advantage}}
    \textcolor{normalpurple}{\FOR{each response $y_i$}
        \STATE {Obtain per-turn judgments $\{\ell_{i,t}\}_{t=1}^{T} = C_\phi(q, y_i, o_{\text{gold}})$}
        \STATE {Map to binary rewards $r^a_{i,t}$ by Eq.~\ref{Eq:turn-reward}}
        \STATE {Compute turn-level advanteage $\{A^a_{i,t}\}_{t=1}^{T}$ using Eq.~\ref{Eq:turn-advantage}}
    \ENDFOR}
    \STATE \
    \STATE \textit{\textcolor{gray}{\# Hybrid Advantage}}
    \textcolor{normalpurple}{\STATE Compute Hybrid Advantage $\widetilde{A}_{i,t}$ via Eq.~\ref{Eq:advantage}.}
\ENDFOR

\STATE \
    \STATE \textit{\textcolor{gray}{\# Policy Optimization}}
    \FOR{step $j \leftarrow 1$ to $M$ do}
        \STATE Update $\pi_\theta$ with Eq.~\ref{Eq:criticsearch_loss}, 
    where per-token loss $\widetilde{\mathcal{L}}_{i,t}$ is computed by Eq.~\ref{Eq:loss}
\ENDFOR

\RETURN $\pi_\theta$
\end{algorithmic}
\label{app:algo}
\end{algorithm}

\onecolumn
\newpage
\section{Template}

\begin{promptbox}[label=box: reward-prompt]{Reward Prompt}
\label{reward_prompt}
Please analyze whether each search process is good or bad, and include the final evaluation result within a <score>...</score> tag. If it is good, output 1; if it is bad, output 0. For example, if there are two actions, output <score>1, 1</score>; if there are four actions, output <score>1, 0, 0, 1</score>.
When making judgments, strictly follow these rules:\\
\hspace*{2em}1. If the "search" action contributes to the final answer or provides partially useful information, it is good.\\
\hspace*{2em}2. If the "search" action provides redundant information, or repeats a search that has already been done, it is bad.\\
\hspace*{2em}3. If the "search" action points to the wrong search direction or misleading information, it is bad.\\
\hspace*{2em}4. If the "search" action results in incorrect information due to unclear expression, it is bad.\\
\hspace*{2em}5. Do not evaluate the "answer" actions, only evaluate the "search" actions.\\
\hspace*{2em}6. If there are no search actions in the entire trajectory, only return <score></score>.\\
\hspace*{2em}7. Only evaluate valid "search" actions (queries that contain <search>...</search> and receive <information>...</information> feedback afterward).\\
\hspace*{2em}8. The final number of scores must match the number of "search" actions.\\
\hspace*{2em}9. Ultimately, first provide a detailed analysis of the search process, then enclose the evaluation results within a <score>...</score> tag.\\

\#\#\#\#\#\#\#\#\#\#\#\#\#\#\#\#\#\#\#\#\#\#\#\#\\
\#\#\#\#\#\#\#\#\#\#\#\#\#\#\#\#\#\#\#\#\#\#\#\#\\
Golden answers: \{Ground truth\}\\
Extracted answer: \{Answer\}\\
Solution string: \{Solution Trajectory\}\\
\#\#\#\#\#\#\#\#\#\#\#\#\#\#\#\#\#\#\#\#\#\#\#\#\\
\#\#\#\#\#\#\#\#\#\#\#\#\#\#\#\#\#\#\#\#\#\#\#\#\\
\\
Please conduct a detailed analysis.\\
The final number of scores should be consistent with the number of search actions.
\end{promptbox}

\end{document}